\documentclass[sigconf,authorversion=true,nonacm=true]{acmart}
\usepackage{enumitem}

\usepackage{booktabs} 

\usepackage{csquotes}
\usepackage{amsmath}
\usepackage{bm}
\usepackage{outlines}

\usepackage{etoolbox}
  \robustify\bfseries

\usepackage{siunitx}
\usepackage{graphicx}

\usepackage{caption}
\usepackage{subcaption}

\usepackage{pifont}
\usepackage{array}
\newcolumntype{P}[1]{>{\centering\arraybackslash}m{#1}}
\newcolumntype{L}[1]{>{\arraybackslash}m{#1}}

\usepackage{xspace}

\newcommand\abs[1]{| #1 |}
\newcommand{\norm}[1]{\left\lVert#1\right\rVert}

\usepackage{url}

\makeatletter
\renewcommand{\fps@figure}{htb}         
\renewcommand{\fps@table}{htb}         
\makeatother 

\usepackage{bbm}
\usepackage[ruled]{algorithm2e}

\usepackage{mathtools}

\usepackage{mleftright}

\DeclarePairedDelimiterX{\infdivx}[2]{[}{]}{%
	#1\;\delimsize\|\;#2%
}

\copyrightyear{2022} 
\acmYear{2022} 
\setcopyright{acmcopyright}\acmConference[WWW '22 Companion]{Companion Proceedings of the Web Conference 2022}{April 25--29, 2022}{Virtual Event, Lyon, France}
\acmBooktitle{Companion Proceedings of the Web Conference 2022 (WWW '22 Companion), April 25--29, 2022, Virtual Event, Lyon, France}
\acmPrice{}
\acmDOI{10.1145/3487553.3524264}
\acmISBN{}

\begin{document}

\title[Web Mining for Electric Vehicle Charging Stations]{Web Mining to Inform Locations of Charging Stations for Electric Vehicles}


\author{Philipp Hummler}
\affiliation{%
	\institution{ETH Zurich}
	\city{Zurich}
	\country{Switzerland}
}
\email{philipp.hummler@me.com}

\author{Christof Naumzik}
\affiliation{%
	\institution{ETH Zurich}
	\city{Zurich}
	\country{Switzerland}
}
\email{cnaumzik@ethz.ch}

\author{Stefan Feuerriegel}
\affiliation{%
	\institution{LMU Munich}
	\city{Munich}
	\country{Germany}
}
\email{feuerriegel@lmu.de}

%
\renewcommand{\shortauthors}{Hummler, Naumzik and Feuerriegel}

\begin{abstract}
The availability of charging stations is an important factor for promoting electric vehicles~(EVs) as a carbon-friendly way of transportation. Hence, for city planners, the crucial question is where to place charging stations so that they reach a large utilization. Here, we hypothesize that the utilization of EV charging stations is driven by the proximity to points-of-interest~(POIs), as EV owners have a certain limited willingness to walk between charging stations and POIs. To address our research question, we propose the use of \emph{web mining}: we characterize the influence of different POIs from \emph{OpenStreetMap} on the utilization of charging stations. For this, we present a tailored \emph{interpretable} model that takes into account the full spatial distributions of both the POIs and the charging stations. This allows us then to estimate the distance and magnitude of the influence of different POI types. We evaluate our model with data from approx. 300 charging stations and 4,000 POIs in Amsterdam, Netherlands. Our model achieves a superior performance over state-of-the-art baselines and, on top of that, is able to offer an unmatched level of interpretability. To the best of our knowledge, no previous paper has quantified the POI influence on charging station utilization from real-world usage data by estimating the spatial proximity in which POIs are relevant. As such, our findings help city planners in identifying effective locations for charging stations. 
\end{abstract}

%
%



\maketitle

\section{Introduction}
\label{sec:introduction} 

Electric vehicles~(EVs) are promoted across large parts of the world as a sustainable, carbon-friendly way of transportation \cite{Needell.2016}. For example, the Netherlands, Denmark, Iceland, and Norway plan policies that will phase out traditional vehicles and enforce sustainable alternatives in form of EVs \cite{Wappelhorst.2021}. 

Crucial for a widespread adoption of EVs among the wider public is the availability of charging stations \cite{Karolemeas.2021,Lin.2011,Thiel.2012}. Currently, many countries are spending massive efforts in building up large-scale infrastructure. For example, the European Union aims to increase the number of charging stations from around 300,000 as of October 2021 to one million until 2025. Hence, this raises the question among both public and private sectors of which locations characterize popular EV charging stations.     

In this paper, we hypothesize that the \textbf{distance} between an EV charging station and points-of-interests~(POIs) is an important determinant explaining the success of EV charging stations. For example, it is likely that EV owners may walk from an EV charging station to a restaurant for only a small distance. Conversely, the distance may be different when EV owners go to a store. To this end, estimating the distance with which POIs influence the success of EV charging stations is of direct importance for city planners: it would directly inform in which proximity around POIs new EV charging stations may need to be located. We propose to address this via web mining of POIs.    

\textbf{Research Question~(RQ):} \emph{Within what distance do POIs contribute to the utilization of EV charging stations?}

\section{Methods}
\label{sec:methods}

\subsection{Problem description}

Our objective is to model and estimate the influence of POIs on the utilization of EV charging stations. To address our research question, we explicitly demand a model that allows for direct \textbf{interpretability} with regard to the POI influence. Specifically, we seek to understand the spatial proximity in which POIs attribute to the popularity of charging stations.  

\underline{EV charging stations:} Let $N$ denote the number of charging stations. Each comes with a location $s_i \in \mathbb{R}^2$, $i = 1, \ldots, N$ and has a utilization $y_i \in [ 0\,\%, 100\,\%]$, $i = 1, \ldots, N$. Each charging station is further characterized by some charger-specific covariates (e.g., information about the neighborhood or the traffic volume). We denote the charger covariates by $x_i \in \mathbb{R}^K$, $i = 1, \ldots, N$ and assume that they could influence the utilization $y_i$ of the corresponding charging station. 

\underline{POI data:} Let $\mathcal{P}$ refer to the set of POIs. Each POI $\rho_j \in \mathcal{P}$ has a location $\omega_j  \in \mathbb{R}^2$ and a discrete type $\gamma_j \in \Gamma$ (e.g., restaurant, public store, public transportation). We write the POI as a tuple $\rho_j = (\omega_j, \gamma_j)$. 

Based on the above, we then estimate a model 
$y_i \sim p\left(f(\bm{x}_i, s_i, \mathcal{P}) \right)$, for $i = 1, \ldots, N$.
Of note, both EV charging stations and POIs have a spatial distribution that must be carefully modeled. As such, we require a spatial model that accounts for the different locations of both. Our model should further consider that each POI (of type $\gamma$) only influences the utilization within a radius $\theta_\gamma$ (``walking distance''). 

\subsection{Model specification}
\label{sec:poi_model}

\underline{Components:} We standardize $y_i$ and model it using a Gaussian likelihood.\footnote{The output of our model parameterizes the mean and the standard deviation is estimated during the training, i.e., $y_i \sim p\left(f(\bm{x}_i, s_i, \mathcal{P}); \sigma \right)$.} As in \cite{Naumzik.2020}, our model consists of three separate components: (1)~a charger influence, (2)~a POI influence, and (3)~spatial heterogeneity.\footnote{Code and data is available from \url{https://github.com/philipphu/poi-ev-charging-stations}.}
\begin{enumerate}[leftmargin=0.6cm]
\item The \textbf{charger influence} is captured by the term $g(\bm{x}_i; \bm{\theta}_x)$, where $\bm{x}_i$ are the charger-specific covariates and $\bm{\theta}_x$ parameters of the function $\bm{g}$. We later set $\bm{g}$ to a neural network (and linear functions in the sensitivity analysis). 
\item The \textbf{POI influence} of a single POI $\rho = (\omega, \gamma)$ is modeled through a scaling factor $\alpha_{\rho}$ (the magnitude of the influence) and a kernel function $k_\gamma(s_i, \omega; \bm{\theta}_{\gamma})$ with parameters $\bm{\theta}_\gamma$. The kernel function accounts for the distance $\norm{s_i - \omega}$ between a charging station at $s_i$ and a POI at $\omega$. Importantly, the scaling factor $\alpha_{\rho}$ is specific to each POI $\rho$ and the kernel function $k_\gamma$ is specific to the type $\gamma$ of the POI. 
\item The \textbf{spatial heterogeneity} $h_0(s_i; \bm{\theta}_{0})$ should capture all remaining variation within cities beyond charger/POI influence (e.g., if a certain city area is more popular). For city planners, this gives the baseline utilization in the city. Following \cite{Banerjee.2014}, the spatial heterogeneity was estimated via a latent zero-mean Gaussian process $\mathcal{GP}(\bm{0}, k_0(\cdot, \cdot; \bm{\theta}_0))$ with a kernel function $k_0$ (here: Mat\'{e}rn 3/2 kernel\footnote{A Mat\'{e}rn 3/2 kernel is preferred in spatial modeling as the infinitely differentiability of the Gaussian kernel is considered unrealistic for physical processes \cite{Stein.1999}.}) and parameters $\bm{\theta}_0$. 
\end{enumerate}

\noindent
\underline{Regression:} Combining the previous components and denoting $\bm{\theta}=\left(\bm{\theta}_x,\bm{\theta}_{\gamma},\bm{\theta}_0\right)$, we yield
\begin{equation*}
f(\bm{x}_i, s_i, \mathcal{P};\bm{\theta}) = 
	\underbrace{
		g(\bm{x}_i; \bm{\theta}_x)
	}_{
		\substack{\text{charger}\\ \text{influence}}
	} + 
	\underbrace{
		\sum_{\rho = (\omega, \gamma) \in \mathcal{P}} \alpha_{\rho} \, k_\gamma(s_i, \omega; \bm{\theta}_{\gamma})
	}_{
		\substack{\text{POI}\\ \text{influence}}
	} + 
	\underbrace{
		h_0(s_i; \bm{\theta}_{0})
	}_{
		\substack{\text{spatial}\\ \text{heterog.}}
	}.
\end{equation*}

\noindent
\underline{Kernel:} In our model, the kernel $k_\gamma$ is responsible for capturing the POI influence by distance. To answer our research question, we use a {ReLU} (rectified linear unit) kernel. It leads to a linear effect up to a certain maximum distance $\theta_\gamma$. Beyond the distance $\theta_\gamma$, the influence of the POI is set to zero, i.e., $k_\gamma(s, \omega;\theta_{\gamma}) = \mathrm{ReLU}\left(1-\frac{\|s- \omega\|}{\theta_{\gamma}}\right)$.

\subsection{Interpretability}
\label{sec:interpretability}

Our POI model allows for direct interpretability to answer our research question. Formally, our model returns two quantities to understand the influence of a certain POI on the utilization of EV charging stations:
\begin{enumerate}[leftmargin=0.6cm]
\item \textbf{Distance.} The ReLU kernel defines a maximum distance (which we refer to as ``cut-off distance'' or ``walking distance'') around POIs within which the POI influence is non-zero. This is provided in form of a radius $\theta_{\gamma}$ (e.g., a POI may only influence the popularity of an EV charging station within 100\,m or 200\,m as EV owners typically do not walk any farther). Crucially, the radius is directly learned from observational data. Moreover, the radius is specific to each POI type $\gamma$. For example, there may be a longer walking distance for POI type ``store'', than for POI type ``restaurant''. 
\item \textbf{Magnitude.} The overall magnitude of the POI influence is determined by the scaling factor $\alpha_{\rho}$. The scaling factor is allowed to vary across POIs, and we thus have an estimate for the influence of each POI. We can also compare the influence of different POI types by computing the average effect size of all POIs of a specific type $\gamma$ via 
$
	\bar{\alpha}_\gamma = \frac{1}{|\mathcal{P}_\gamma|}\sum_{\rho\in\mathcal{P}_\gamma}{|\alpha_{\rho}|} $,
 where $\mathcal{P}_\gamma$ is the set of POIs of type $\gamma$.
\end{enumerate}

\subsection{Estimation}

We estimate the model via the following efficient learning algorithm. Specifically, by adapting \cite{Naumzik.2020}, we yield a tailored sparse variational inference for our problem.
We use the combined influence of all POIs of type $\gamma$, i.e.,
\begin{equation*}
    h_\gamma(s_i, \bm{P}_\gamma; \bm{\theta}_{\gamma}) = \sum_{\rho = (\omega, \gamma)\in \bm{P}_\gamma} \alpha_{\rho} \, k_{\gamma}(s_i, \omega;\bm{\theta}_{\gamma})
\end{equation*}
to write our model as 
\begin{equation*}
    f(\bm{x}_i, s_i,  \mathcal{P}; \bm{\theta}) = g(\bm{x}_i; \bm{\theta_x}) + \sum_{\gamma \in \Gamma} h_\gamma(s_i, \bm{P}_\gamma; \bm{\theta_\gamma}) + h_0(s_i; \bm{\theta}_{0}), 
\end{equation*}
for $i = 1, \ldots, N$.
As shown in \cite{Naumzik.2020}, $h_\gamma$ is given by a zero-mean Gaussian process $\mathcal{GP}(\bm{0}, \tilde{k}_{\gamma}(\cdot, \cdot;\mathbf\theta_{\gamma}))$.\footnote{$\tilde{k}_{\gamma}(s,s';\bm{\theta}_{\gamma}) = \mathbb{E}\left[h_\gamma(s)\,h_\gamma(s')\right]$. See also \cite{Naumzik.2020}.}

Let $\Gamma_0 = \Gamma\cup\{0\}$ be the index set for the Gaussian processes $\bm{h}_0$ and $\bm{h}_{\gamma}$.
We have the likelihood $p(\bm{y} \mid \bm{f})$ for our model. Its form is chosen in accordance with the distribution of our target variable $y_i$.
We derive the so-called evidence lower bound~(ELBO), which gives a lower bound for the marginal likelihood $p(y)$, to find a variational Gaussian approximation $q(\bm{h}_{\gamma})$ of the true posterior distribution $p(\bm{h}_{\gamma} \mid \bm{y})$. 
We further improve the scalability through a sparse approximation using an extension of the inducing point method from \cite{Hensman.2015}. That is, we sample $M$ inducing points from the locations of the charging stations and denote with $\bm{u}_{\gamma}$ the output of the Gaussian process $\bm{h}_{\gamma}$ at the inducing points. The ELBO with inducing points for additive latent Gaussian processes is then 
\begin{equation}
\begin{split}
	\log p(\bm{y})&\geq\mathbb{E}_{q(\bm{h}_0)\,q(\bm{h}_{\gamma_1}) \cdots q(\bm{h}_{\gamma_{\abs{\Gamma}}})}\left[\log p(\bm{y}\vert\bm{h}_0,\bm{h}_{\gamma_1},...,\bm{h}_{\gamma_{\abs{\Gamma}}})\right] \\
	&\phantom{=} -\sum_{\gamma\in\Gamma_0}\textup{KL}\infdivx[\big]{q(\bm{u}_{\gamma})}{p(\bm{u}_{\gamma})}.
\end{split}
\end{equation}
The training objective is then to maximize this ELBO.
Finally, we generate predictions for $\bm{h}_{\gamma}$ at a new location $s^\ast$ via
\begin{equation}
q(\bm{h}_\gamma^\ast) = \int p(\bm{h}_\gamma^\ast \mid \bm{u}_\gamma) \, q(\bm{u}_\gamma) \, \mathrm{d}\bm{u}_\gamma.
\end{equation}

\noindent
Of note, hyperparameters in the above POI model are absent. Rather, all parameters can be directly estimated from observational data. The training of our POI model took less than two minutes using a 12-core CPU. For the evaluation, we split the data into a training and a test set using an 80:20 ratio.

\subsection{Baselines}
\label{sec:baselines}

We compare our model against several state-of-the-art baselines for POI modeling, which combine (1)~feature engineering and (2)~a prediction model. Here, our feature engineering is analogous to earlier research \cite{Wagner.2016,Xiao.2017,Kadar.2019,Karamshuk.2013,Wang.2015,Willing.2017}:
\begin{enumerate}[leftmargin=0.6cm]
\item \emph{Distance-based POI features} \cite{Wagner.2016,Xiao.2017}: For each $y_i$ and each POI type $\gamma$, we compute the distance between the location $s_i$ of $y_i$ and the closest POI of type $\gamma$. Hence, this provides the shortest distance to the next POI of a given type.
\item \emph{Density-based POI features} \cite{Kadar.2018b,Kadar.2019,Karamshuk.2013,Wang.2015,Willing.2017}: For each observation $y_i$ at location $s_i$ and each POI type $\gamma$, we count the number of POIs of type $\gamma$ at locations $\omega_j$ within a given distance $D_{\max}^\gamma$ to the observation; i.e., $\left|\left\{\rho_j \in \mathcal{P} : \left\|s_i - \omega_j\right\| < D_{\max}^\gamma \right\}\right|$. Hence, this captures the relative density of POIs within a spatial area. Here, $D_{\max}^\gamma$ is tuned via grid search.
\end{enumerate}

\noindent
Using the above feature engineering, we then use the following \textbf{spatial} baselines: geographically weighted regression (\textbf{GWR}) \cite{Brunsdon.1996}, \textbf{linear kriging} \cite{Banerjee.2014}, \textbf{RF kriging} with random forest \cite{Naumzik.2020}, and a deep \textbf{neural network} \cite{Naumzik.2020}. 

\section{Data}
\label{sec:data}

\textbf{EV charging stations:} Our data comprises $N = 287$ charging stations from Amsterdam \cite{Brandt.2021}. For them, we generated a dataset consisting of location $s_i$ and average utilization $y_i$.\footnote{Specifically, the utilization was obtained by repeatedly querying the public API \url{https://www.nuon.nl/ev/publiek/kaartoplaadpunten.do} every minute about the real-time usage status. Altogether, this resulted in $\sim$150 million observations for an eight-month period (July 01, 2013 through January 31, 2014). These were then aggregated into an overall utilization per charging station. For charging stations with multiple outlets, the mean utilization of all outlets was used.} 

For each charging station, additional charger covariates $x_i$ were retrieved: the population density of the neighborhood; the average income per person in the neighborhood (log-transformed); the car density of the neighbourhood; and a binary dummy whether a major road is within 250\,m. Here, neighborhood refers to one of the 15 official boroughs for administrative purposes. 

\textbf{POI data:} POIs were obtained from OpenStreetMap.\footnote{\url{https://www.openstreetmap.org}} For better interpretability, we choose the following POI types $\gamma$: (i)~restaurants, (ii)~stores (i.e., clothing stores, department stores, grocery stores, supermarkets, and shopping malls), (iii)~education (i.e., schools and universities), and (iv)~public transportation (i.e., stations for bus, metro, and train). The latter is relevant for Amsterdam as it is common for employees from outside of Amsterdam to use so-called park-and-ride offers to travel to the city center. As a result, our dataset contains 4,036 POIs.

\section{Results}
\label{sec:results}

\textbf{Overall performance: }
Table~\ref{tbl:results_chargingpoints} compares the different models across (1)~the out-of-sample root mean squared error (RMSE) in predicting $y_i$ and (2)~the log-likelihood of the corresponding model. For the baselines, we report variants with and without POI features. The additional POI information improved the performance of GWR and linear kriging, while the opposite is true for RF kriging and the neural network. Our POI model outperforms all baselines. 

\begin{table}[htb]
\centering
\scriptsize
\setlength\tabcolsep{2pt}
\begin{tabular}{lSS}
\toprule
Model [feature engineering]& \multicolumn{1}{c}{RMSE} & \multicolumn{1}{c}{Log-lik.}\\
\midrule
GWR [none] &1.164 & -86.396 \\
GWR [both] &1.131 & -84.793 \\
Linear kriging [none] &1.116 & -84.102 \\
Linear kriging [both] &1.094 & -82.979 \\
RF kriging [none] &1.148 & -85.638 \\
RF kriging [both] &1.162 & -86.285 \\
Neural network [none] &1.125 & -84.501 \\
Neural network [both] &1.200 & -88.102 \\
\midrule
POI model (ours) &\bfseries 1.066 & \bfseries -81.561 \\
\bottomrule
\multicolumn{3}{l}{Note: Best value per column is in bold.}
\end{tabular}
\caption{Out-of-sample performance of the different models.}
\label{tbl:results_chargingpoints}
\end{table}

\textbf{Interpretation of POI influence: }
We now proceed by answering our research question. Here, we remind that a core strength of our POI model is its ability to estimate the parameters describing the influence of different POI types. Specifically, we can interpret both (1)~the maximimum distance $\theta_\gamma$ (``radius'') and (2)~the magnitude of the effect, $\overline{\alpha}_\gamma$. See Sec.~\ref{sec:interpretability} for details. The estimation results are in Table~\ref{tbl:effects_chargingpoints}. 

We arrive at the following findings in response to our research question: (1)~The influence of POIs varies across different types. The largest distance $\theta_\gamma$ for a non-zero influence is found for POIs of type ``education''. Here, the distance is approx. twice as large as the corresponding distance for restaurants, stores, and public transportation stops. This implies that, on average, EV owners are more willing to walk longer between charging stations and education-related POIs (i.e., schools or universities). For example, our model estimates the walking distance to 297\,m for restaurants and to 643\,m for buildings for education. (2)~Schools and universities have the largest overall effect size in attracting a high utilization of EV charging stations.  

\begin{table}[htbp]
\centering
\scriptsize
\setlength\tabcolsep{2pt}
\begin{tabular}{l SSS}
\toprule
&\multicolumn{1}{c}{Distance}&\multicolumn{2}{c}{Magnitude}\\
\cmidrule(lr){2-2}
\cmidrule(lr){3-4}
POI type $(\gamma)$ & {Cut-off $\theta_{\gamma}$ [in km]} & {Average effect $\bar{\alpha}_\gamma$} & {SD}\\		
\midrule
Restaurant & 0.297 & 0.002 & 0.007\\
Store & 0.279 & 0.001 & 0.004\\
Education & 0.643 & 0.031 & 0.055\\
Public transportation & 0.351 & 0.003 & 0.013\\
\bottomrule
\multicolumn{1}{l}{SD: standard deviation}
\end{tabular}
\caption{Estimated parameters of the POI influence.}
\label{tbl:effects_chargingpoints}
\end{table}

\begin{figure}[htbp]
	\begin{subfigure}{.23\textwidth}
		\centering
		\includegraphics[width=0.9\linewidth]{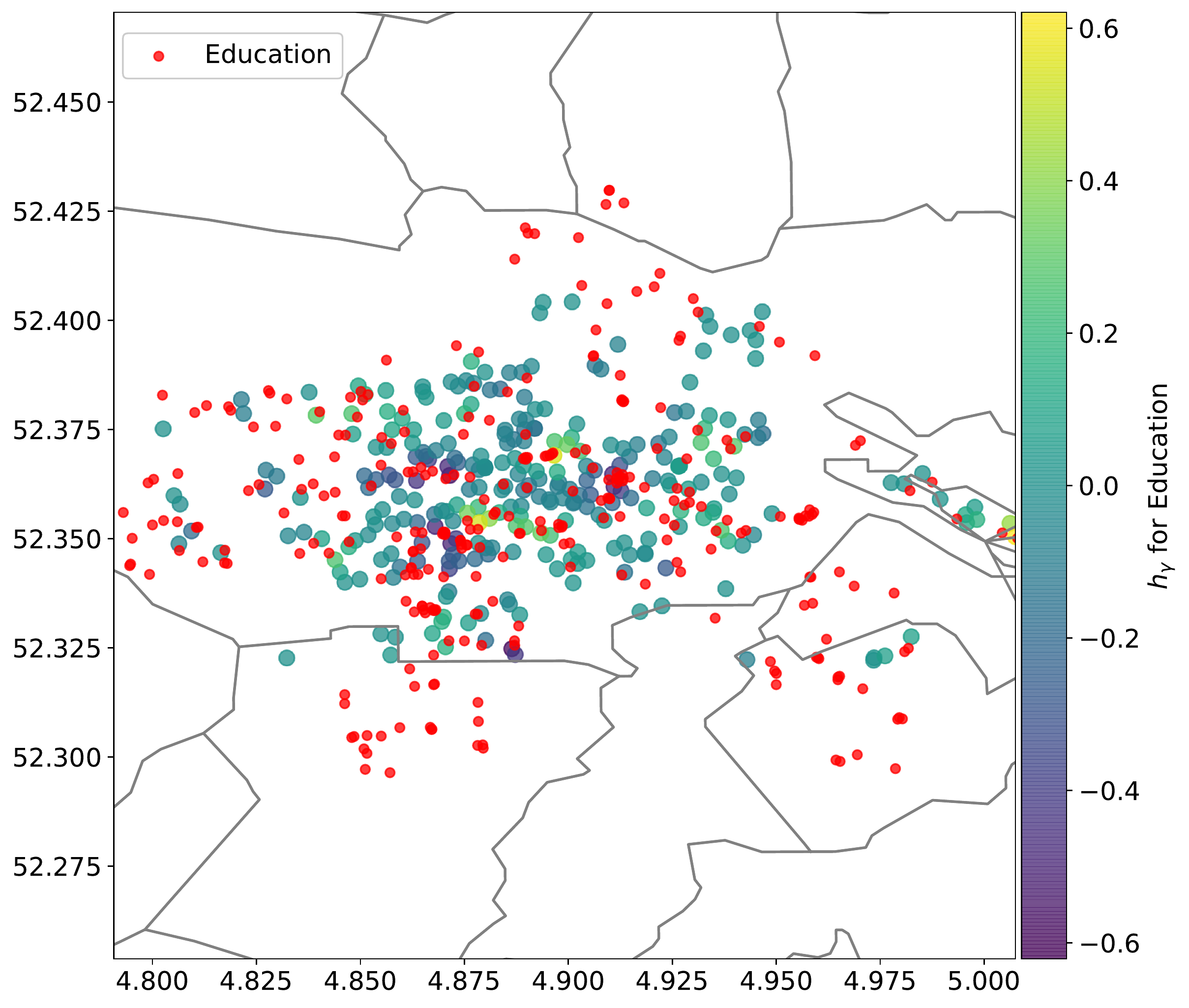}
	\end{subfigure}
	\begin{subfigure}{.23\textwidth}
		\centering
		\includegraphics[width=0.9\linewidth]{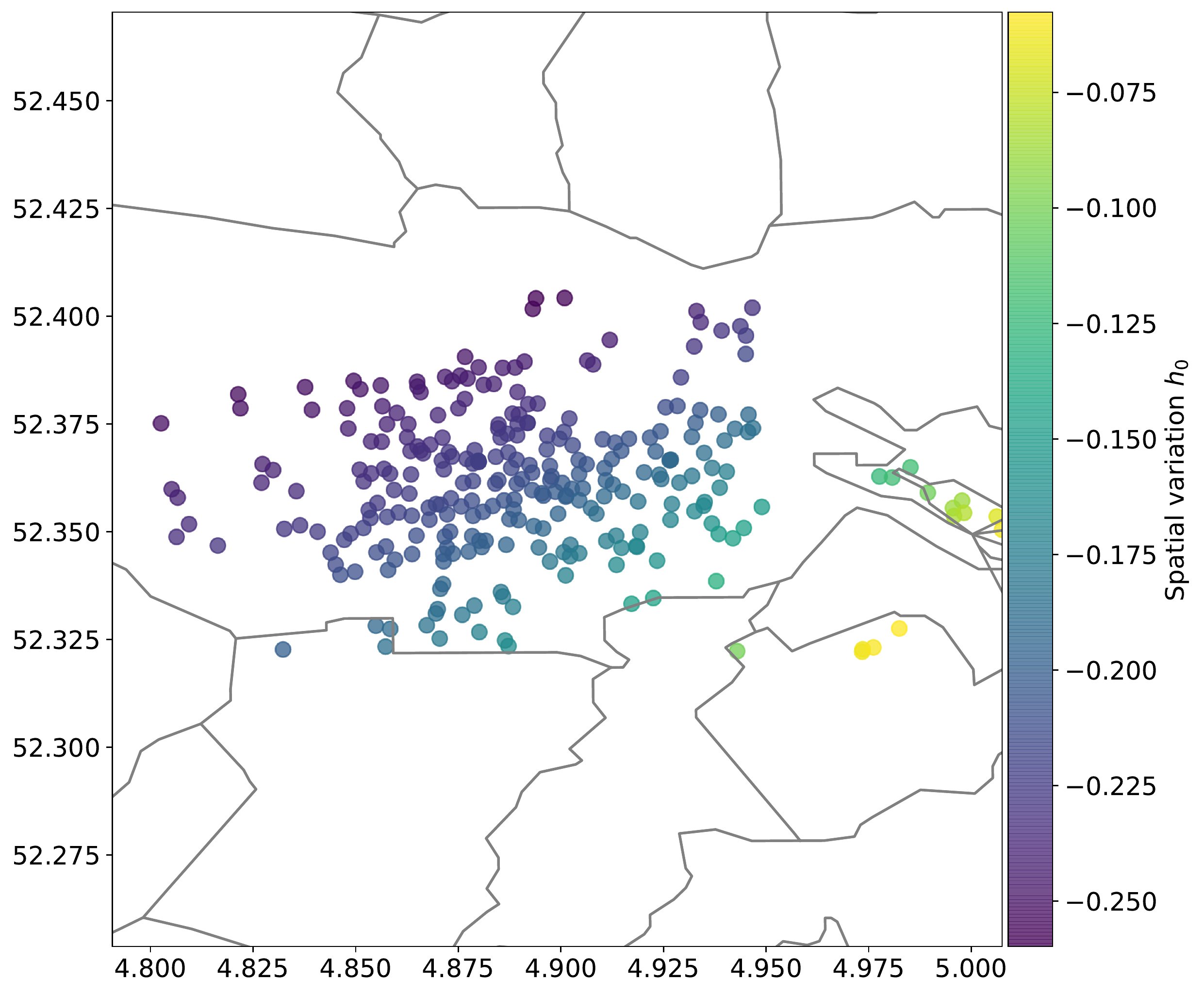}
	\end{subfigure}
	\caption{\emph{Left:} Recovered latent POI influence $\bm{h}_\gamma$ for POI type ``education'' on charging station utilization (shown by color gradient). All POIs of type ``education'' are further shown in red. \emph{Right:} Remaining spatial heterogeneity $h_0$ with baseline utilization.}
	\label{fig:spatial_variation}
\end{figure}

Finally, we also plot the remaining spatial heterogeneity $h_0$ in Fig.~\ref{fig:spatial_variation}. Here, for instance, we observe a strong negative influence in the northwest of Amsterdam that gradually becomes weaker when moving towards the southeastern part of the city. This points to neighborhoods where the baseline utilization is high.

\textbf{Sensitivity analysis: }
Our model allows for different degrees of freedom: one can use (1)~different functions $\bm{g}$ to model the charger influence (e.g., neural networks as a non-linear alternative) and (2)~different kernel functions. In addition to the ReLU kernel, we also tested a Gaussian kernel function. The {Gaussian} kernel yields a decay following a normal distribution as the distance increases, i.e., $k_\gamma(s, \omega;\theta_{\gamma}) = \exp\left(-\frac{\|s-\omega\|^2}{2 \, {\theta_{\gamma}}^2}\right)$. Results are in Table~\ref{tbl:sensitivity_analysis_chargingpoints}. The ReLU kernel outperforms the Gaussian kernel. The neural network outperforms the linear choice for $\bm{g}$ at the expense of interpretability.

\begin{table}[htbp]
\centering
\scriptsize
\setlength\tabcolsep{2pt}
\begin{tabular}{lllSScS}
\toprule
{Charger influence $\bm{g}$} & {Kernel $k_\gamma$} & {RMSE} & {Log-lik.}\\
\midrule
Neural network & ReLU & 1.066 & -81.561 \\
Linear & ReLU & 1.088 & -82.720 \\
Neural network & Gaussian & 1.106 & -83.571 \\
Linear & Gaussian & 1.115 & -84.041 \\
\bottomrule
\end{tabular}
\caption{Sensitivity analysis comparing the ouf-of-sample performance across different model specifications.}
\label{tbl:sensitivity_analysis_chargingpoints}
\end{table}


\section{Related Work}
\label{sec:related_work}

POIs provide geo-tagged data that refer to locations, which people may find useful or important \citep{Rae.2012}. Examples of POIs are, e.g., restaurants, stores, universities, and public transport stations. POI data has now become widely available due to web mapping services like Google or OpenStreetMap. 

Previous research has used POI data for different purposes; e.g.: (1)~There are works on POI recommendation where a list of user-specific suggestions for future visits is made (e.g., \cite{Ye.2011}). (2)~There are works inferring POI data from other sources such as web data (e.g., \cite{Rae.2012}). However, both streams (1) and (2) are different from ours as POIs are output and not predictors. (3)~A different stream uses POIs to explain various social phenomena \cite{Cranshaw.2012,Hristova.2016,Taylor.2018,Hidalgo.2020,Noulas.2012,Yuan.2012,Schwabe.2021} including housing prices \cite{Xiao.2017, Fu.2019, Tang.2018} or crime \cite{Kadar.2019, Wang.2016}. Different from them, our aim is to estimate the distance of the POI influence on EV charging stations. 

\section{Discussion}
\label{sec:discussion}

Surveys suggest that the proximity to POIs is one of the success factors of charging station utilization \cite{Karolemeas.2021}. However, this was obtained via surveys, and not via observational inference through web mining of real-world data. To this end, our results find a profound influence of POIs on the utilization of an EV charging station that, further, strongly depends on the distance. Evidently, people walk between educational facilities and charging stations roughly twice as long as for stores, restaurants, and public transportation stops. 

For the adoption of EVs, careful planning of the charging network is necessary. The utilization of charging stations is highly dependent on the convenience for customers, that is, the walking distance to surrounding POIs is acceptable for EV owners. Here, our web mining approach helps city planners to determine suitable locations for charging stations by indicating within which distance to certain POIs they are expected to reach a high utilization.

\bibliographystyle{ACM-Reference-Format-no-doi}
  \newcommand{\dq}{"}
\bibliography{literature}

\end{document}